\documentclass{article}
\IfFileExists{iclr2027_conference.sty}{%
  \usepackage{iclr2027_conference,times}%
}{%
  \usepackage[margin=1in]{geometry}
  \usepackage{natbib}
  \PackageWarningNoLine{cb-agent}{ICLR style unavailable; using article preview layout}
}
\usepackage[T1]{fontenc}
\usepackage[utf8]{inputenc}
\usepackage{amsmath,amssymb,amsthm,bm}
\usepackage{booktabs}
\usepackage{longtable}
\usepackage{multirow}
\usepackage{graphicx}
\usepackage{algpseudocode}
\usepackage{hyperref}
\usepackage{url}

\newcounter{cbalgorithm}

\title{CB-Agent: Concept-Based In-Context Learning for Sequential Action Selection}
\author{
  Qingyue Zhang\thanks{Equal contribution.}
  \quad
  Huaze Tang\textsuperscript{*}
  \quad
  Minyu Zhang
  \quad
  Yuxiang Yan
  \quad
  Wenbo Ding\thanks{Corresponding authors:
    Wenbo Ding (\href{mailto:ding.wenbo@sz.tsinghua.edu.cn}
    {\texttt{ding.wenbo@sz.tsinghua.edu.cn}});
    Shao-Lun Huang (\href{mailto:twn2gold@gmail.com}
    {\texttt{twn2gold@gmail.com}}).}
  \quad
  Shao-Lun Huang\textsuperscript{\dag}
  \\[1.5ex]
  {\normalfont Tsinghua Shenzhen International Graduate School,}
  \\
  {\normalfont Tsinghua University, Shenzhen, China}
}
\iclrfinalcopy
\begin{document}
\maketitle

\begin{abstract}
In-context learning (ICL) allows large language models to use
demonstrations without updating their parameters. For interactive agents, demonstrations are trajectories of reasoning, actions, and observations that illustrate how to complete a task. Reusing demonstrations in new situations is challenging: directly inputting full trajectories consumes limited prompt space and constrains the overall effectiveness, while natural-language summaries require the agent to translate general guidance into specific action choices. Therefore, we study whether expert action choices
can directly supervise a reusable scoring rule.
We propose CB-Agent, which extends concept-based ICL to sequential action selection. We extract environment-action decisions from expert trajectories and fit a concept vector by maximum likelihood to model the conditional distribution over candidate actions given the context. This model combines the concept with frozen language-model features and priors. During ReAct interaction, the agent uses this distribution to select environment actions without including demonstrations in the action-scoring context.
Across TextWorld, ALFWorld, and ScienceWorld, CB-Agent achieves the
highest cross-benchmark average among the evaluated methods at all three
tested model sizes and ranks first on TextWorld and ScienceWorld at
each size. ALFWorld experiments support the learned concept's contribution
beyond prior-only scoring and show further gains when combined with Reflexion.
Code is available at \url{https://anonymous.4open.science/r/CB-Agent-DFF9/}.
\end{abstract}

\section{Introduction}
\label{sec:introduction}
In-context learning (ICL) enables a large language model (LLM) to perform
a task using examples in its input, without updating its parameters
\citep{brown2020language}.
These examples, called demonstrations, typically pair an input with a
desired output. For an interactive agent, demonstrations instead record
trajectories of actions and observations. Few-shot ReAct includes these demonstration trajectories directly in the
prompt to illustrate how reasoning and actions can be combined as new
observations arrive \citep{yao2023react}.
The agent must use this experience throughout execution: each decision
depends on the interaction so far and can affect the steps that follow.
The goal is therefore to reuse demonstration knowledge across changing
contexts to complete a new task.

Existing approaches improve how language models use experience through
demonstration selection, task organization, and feedback.
Fine-tuning offers another route, but collecting interaction data or
human annotations and optimizing model parameters can be costly
\citep{nakano2021webgpt}.
Within ICL, retrieval-based selection identifies demonstrations relevant
to a query input \citep{liu2022goodexamples}.
For interactive tasks, StateAct maintains explicit goals and state
information to guide decisions \citep{rozanov2025stateact}.
AdaPlanner refines its plans in response to environmental feedback
\citep{sun2023adaplanner}.
To reuse accumulated experience, ExpeL extracts natural-language insights
and combines them with retrieved successful trajectories
\citep{zhao2024expel}.
Reflexion instead maintains verbal reflections on feedback to guide
subsequent attempts \citep{shinn2023reflexion}.
Demonstrations in a prompt nevertheless compete with interaction history
for limited space, while reflection-based retries require additional
model calls and environment interactions. For demonstration-based action
selection, we investigate an alternative: using expert choices to fit a
reusable scoring rule, rather than relying solely on trajectories or
verbal guidance in the action-selection prompt.

The central challenge is how experience should affect each action choice.
A new task need not reproduce any demonstrated sequence, and applying
language-based insights requires interpreting their guidance in the
current situation.
For example, advice to avoid repeated searches must distinguish an
unproductive revisit from returning to a known location to collect a
second object. Expert trajectories also provide direct supervision:
the action selected in each recorded context. This motivates our central
question: \emph{can we estimate a reusable action-selection rule directly
from demonstrated choices, without first expressing that experience as
natural-language rules?} Fitting these choices would explicitly connect
demonstration learning to the comparison of candidate actions.

Numerical task representations provide a foundation for this direction.
Task Vectors and Function Vectors show that demonstrated tasks can be
represented by vectors that influence a model's predictions
\citep{hendel2023taskvectors,todd2024functionvectors}.
More directly, concept-based ICL (CB-ICL) estimates a shared concept
vector over frozen language-model features and provides theoretical
analyses of label prediction for individual inputs \citep{tang2025theoretical}.
Here, a concept is a numerical vector that weights features for
prediction. Adapting this framework to sequential action selection
requires extracting decision examples from trajectories and handling
contexts and candidates that change with interaction. The concept must
express context-dependent preferences, not a fixed ranking of action strings.


We introduce CB-Agent, which converts expert trajectories into decision
examples comprising a pre-action context, candidate environment actions,
and the expert's choice. We model the conditional distribution over
candidate actions given the context using an exponential-family model.
A concept vector weights frozen LLM features to adjust the language-model
prior and is fitted to the expert choices by regularized maximum likelihood.
During ReAct interaction, the fitted model selects the highest-probability
candidate environment action. ReAct continues to use demonstrations for
reasoning and think-or-act routing, while the action scorer uses the current
context without demonstrations. The concept remains fixed during execution,
but action probabilities change with the context and candidates.
This procedure supports both environment-supplied candidates in
fixed-action-space tasks and model-generated candidates in
open-action-space tasks.

Our contributions can be summarized as follows:
\begin{enumerate}
 \item \textbf{Concept-based learning from trajectory decisions.}
 We adapt CB-ICL to estimate a shared concept from expert environment-action
 choices. The conditional model combines the learned feature weights with
 a language-model prior, linking demonstration fitting to candidate scoring
 without updating LLM parameters or requiring verbal experience summaries.

 \item \textbf{Online selection with a reusable concept.}
 We integrate the fitted model into ReAct interaction, using one fixed
 concept with changing contexts and candidates in both fixed and open
 environment-action spaces.

 \item \textbf{Empirical evaluation and analysis.}
 We evaluate Qwen3-4B, Qwen3-8B, and Qwen3-14B on TextWorld, ALFWorld,
 and ScienceWorld. CB-Agent achieves the highest cross-benchmark average
 among the evaluated methods at each model size and ranks first on
 TextWorld and ScienceWorld throughout. An ALFWorld control comparison
 supports the contribution of the learned concept beyond the
 length-adjusted language-model prior. We also examine task-related
 concept structure and evaluate combinations with Reflexion separately.
\end{enumerate}

\section{Background and Problem Setup}
\label{sec:background}
This section reviews ICL (Section~\ref{sec:icl-background}) and
concept-based prediction (Section~\ref{sec:cbicl-background}), then defines
the sequential action-selection problem (Section~\ref{sec:sequential-setup}).

\subsection{In-Context Learning}
\label{sec:icl-background}
In standard few-shot ICL, input--output demonstrations accompany a query
input in the prompt, guiding its prediction without updating language-model
parameters \citep{brown2020language}.

For agents, demonstrations are trajectories of task instructions, actions,
and observations. Few-shot ReAct includes reasoning in these trajectories
to guide interaction \citep{yao2023react}.
Unlike single-output prediction, interaction requires successive decisions
conditioned on the evolving history.

\subsection{Concept-Based In-Context Learning}
\label{sec:cbicl-background}
Concept-based in-context learning (CB-ICL) estimates a shared concept vector
from labeled demonstrations using features from a frozen language model
\citep{tang2025theoretical}.
The estimated vector weights the features of an input and its candidate
labels, allowing demonstrations to inform a common prediction rule.
The published CB-ICL analysis uses a linear representation of conditional
probabilities.

In this work, we adopt an exponential parameterization instead of the
linear form, motivated by local information geometry.
Under weak dependence, where the joint distribution is close
to the product of its marginals in chi-squared divergence, the linear
modal decomposition admits a local exponential representation, up to
higher-order terms \citep{huang2024universal}.
This representation motivates weighting fixed features in log-probability
space. Appendix~\ref{app:exponential-motivation} gives the derivation and
distinguishes this local approximation from the practical concept model.

Let $x$ be an input text, $\mathcal Y$ a finite, nonempty set of candidate
labels, and $y\in\mathcal Y$ a label. Let $m$ be the positive integer
feature dimension, $\bm\phi(x,y)\in\mathbb R^m$ a fixed feature vector,
and $Q_0(y)>0$ a fixed base weight. A concept vector
$\bm\alpha\in\mathbb R^m$ defines the conditional distribution
$P_{\bm\alpha}$ by multiplying each base weight by a feature-dependent
factor and normalizing over labels:
\begin{equation}
 P_{\bm\alpha}(y\mid x,\mathcal Y)=
 \frac{Q_0(y)\exp\{\bm\alpha^\top\bm\phi(x,y)\}}
 {\sum_{b\in\mathcal Y}Q_0(b)\exp\{\bm\alpha^\top\bm\phi(x,b)\}}.
 \label{eq:cbicl-exponential}
\end{equation}
Here $\top$ denotes transpose, and $b$ indexes candidate labels.
The concept is estimated by maximizing the conditional likelihood of
the demonstrated labels and is then shared across inputs.
While prior work on CB-ICL considers label prediction for individual
inputs, we adapt concept-based prediction to sequential action
selection for interactive agents (Section~\ref{sec:method}).

\subsection{Sequential Action Selection and Problem Formulation}
\label{sec:sequential-setup}
Let \(\mathrm{inst}\) denote the task instruction and \(o_0\) the initial observation provided by the environment before any action is taken. If no initial observation is provided, we treat \(o_0\) as empty.
At interaction step $t=0,1,\ldots$, the action string $a_t$ is either
an environment action or a \texttt{Think} action containing reasoning text.
The recorded feedback $o_{t+1}$ is an environment observation for the former
and the fixed response \texttt{OK} for the latter, which leaves the environment
unchanged. Both action types count as interaction steps.
For a positive integer interaction length $T$, a complete trajectory is
\begin{equation}
 \tau=(\mathrm{inst},o_0,a_0,o_1,\ldots,a_{T-1},o_T).
 \label{eq:interaction-trajectory}
\end{equation}
Let $\mathcal D_{\mathrm{traj}}$ denote the collection of expert
trajectories, whose lengths may differ.

At the start of a new task, the agent receives the task instruction
$\mathrm{inst}$ and initial observation $o_0$.
Before selecting action $a_t$, it constructs a text context $x_t$ from
the task instruction, the initial observation, and all preceding actions
and feedback. The agent either generates a \texttt{Think} action or selects
an environment action. At an environment-action step, it selects $a_t$
from a finite, nonempty candidate collection $\mathcal A_t$.
In both cases, it records $o_{t+1}$ and updates the context.
The trajectory is thus generated through interaction rather than
predicted in advance.

Given $\mathcal D_{\mathrm{traj}}$ and a frozen language model, our
objective is to solve new tasks from the same task distribution through
this interaction process, without updating the model parameters.
Success is determined by whether the interaction achieves the goal
specified by $\mathrm{inst}$, rather than by action prediction accuracy alone.
Section~\ref{sec:concept-estimation} describes how expert trajectories
provide the decision examples used to learn the action-selection rule.

\section{CB-Agent: Concept Estimation and Action Selection}
\label{sec:method}
\begin{figure}[t]
 \centering
 \includegraphics[width=\textwidth]{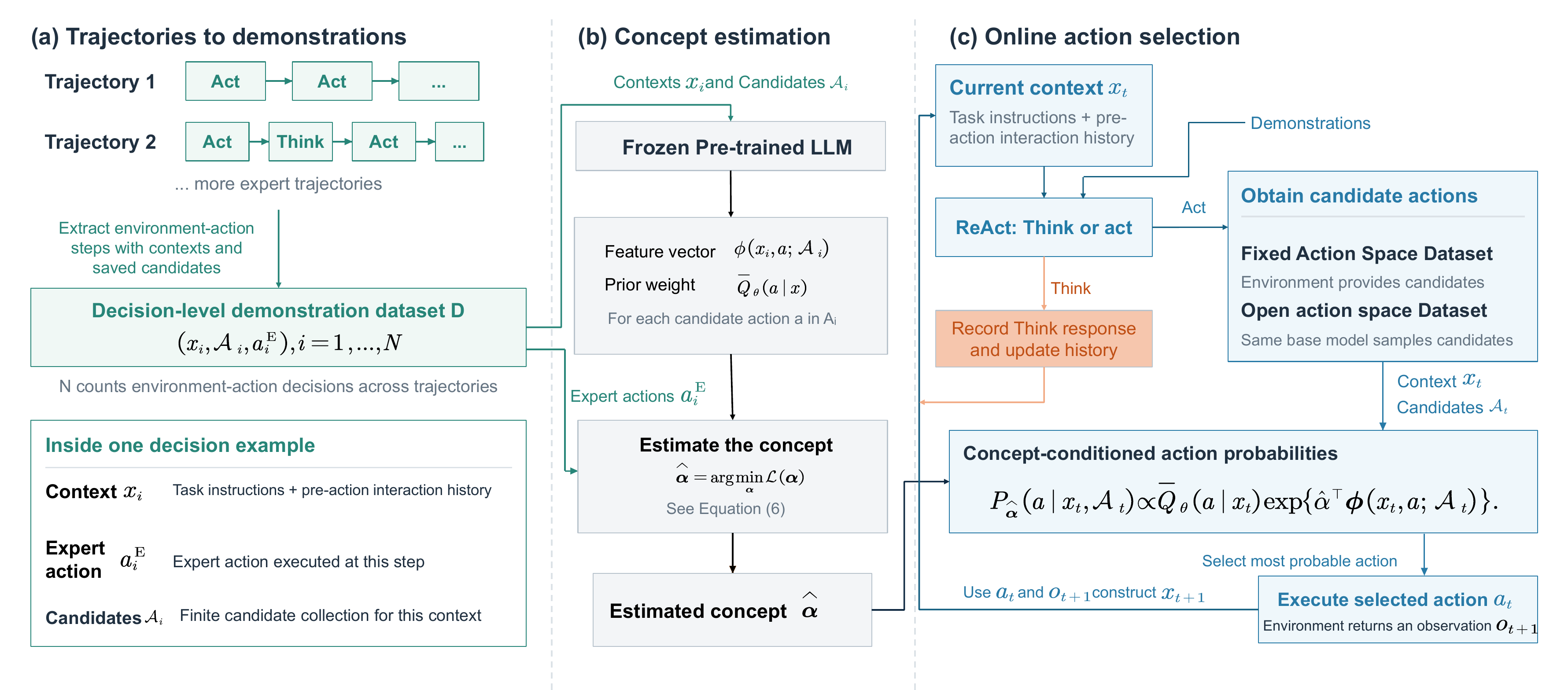}
 \caption{\textbf{Overview of CB-Agent.}
 (a) Only environment-action steps in expert trajectories provide decision
 examples comprising a pre-action context, candidate actions, and the
 expert action. Preceding thoughts remain in the context.
 (b) A frozen language model supplies action features and length-adjusted
 prior weights; a shared concept is estimated by regularized maximum
 likelihood.
 (c) A ReAct response determines whether to think or act. A \texttt{Think}
 response is recorded in the interaction history. On the action branch,
 the agent constructs candidate actions and selects one using the estimated
 concept and prior weights.
 The interaction history is updated after either branch, while the concept
 and language model remain fixed.}
 \label{fig:cb-agent-overview}
\end{figure}

In this section, we present CB-Agent.
Section~\ref{sec:action-model} specifies the conditional action model.
Section~\ref{sec:concept-estimation} constructs decision examples from
expert trajectories and defines the concept-estimation objective.
Finally, Section~\ref{sec:online-selection} describes online action
selection with the fitted concept. Algorithm~\ref{alg:cb-agent}
summarizes the complete procedure.
Figure~\ref{fig:cb-agent-overview} illustrates how environment-action
steps in expert trajectories provide demonstrations for concept estimation
and how the fitted concept guides action selection on the ReAct action branch.

\subsection{Concept-Conditioned Action Probabilities}
\label{sec:action-model}
We instantiate Equation~\eqref{eq:cbicl-exponential} with interaction
contexts as inputs and candidate environment actions as labels. For one environment-action decision,
we omit the step index and write $x$ for the pre-action context,
$\mathcal A$ for the finite, nonempty candidate collection, and
$a\in\mathcal A$ for a candidate environment action. Unlike a fixed-label prediction problem,
both the candidates and their language-model weights can change across
decisions.

Let $\theta$ denote the frozen language-model parameters. Retaining the
feature dimension $m$ and concept vector $\bm\alpha$ from
Section~\ref{sec:cbicl-background}, let
$\bm\phi(x,a;\mathcal A)\in\mathbb R^m$ be the feature vector of action $a$.
We construct it from the model's last-layer representation at the final
action token, with coordinate-wise standardization across candidates.
The features therefore depend on the context and candidate collection.
Appendix~\ref{app:implementation} specifies the construction.

Let $Q_\theta(a\mid x)>0$ be the model probability of the action's token
sequence, $L(a)$ its positive integer token count, and
$\gamma\in[0,1]$ a fixed length-correction exponent.
Following the power-law length normalization discussed by
\citet{wu2016googlesneuralmachinetranslation}, we use the length-adjusted prior weight
$\overline Q_\theta(a\mid x)=Q_\theta(a\mid x)^{1/L(a)^\gamma}$.
Replacing the base weight in Equation~\eqref{eq:cbicl-exponential}
with this context-dependent weight gives the action distribution:
\begin{equation}
 P_{\bm\alpha}(a\mid x,\mathcal A)=
 \frac{\overline Q_\theta(a\mid x)
       \exp\{\bm\alpha^\top\bm\phi(x,a;\mathcal A)\}}
      {\sum_{b\in\mathcal A}\overline Q_\theta(b\mid x)
       \exp\{\bm\alpha^\top\bm\phi(x,b;\mathcal A)\}}.
 \label{eq:conditional-action-model}
\end{equation}
Here $b$ indexes candidate actions. The concept adds the score
$\bm\alpha^\top\bm\phi(x,a;\mathcal A)$ to the prior log-weight.
Let $\bm 0$ denote the zero vector in $\mathbb R^m$.
Setting $\bm\alpha=\bm 0$ recovers the normalized length-adjusted prior.
The same model is used for concept estimation and action selection.

For two candidates $a,b\in\mathcal A$, their log-probability ratio
separates the prior preference from the learned adjustment:
\begin{equation}
 \begin{aligned}
 \log\frac{P_{\bm\alpha}(a\mid x,\mathcal A)}
               {P_{\bm\alpha}(b\mid x,\mathcal A)}
 &=\log\frac{\overline Q_\theta(a\mid x)}{\overline Q_\theta(b\mid x)}+\bm\alpha^\top
 [\bm\phi(x,a;\mathcal A)-\bm\phi(x,b;\mathcal A)].
 \end{aligned}
 \label{eq:relative-preference}
\end{equation}
One concept vector is fitted to the expert trajectories selected for a
task and shared across its environment-action decisions, giving a common
scoring rule. This does not impose a fixed preference over action
strings: their features change with the context and candidates.
The learned adjustment can reinforce or reverse the prior ranking.

This exponential family models expert action preferences; its scope
and limitations are discussed in Appendix~\ref{app:limitations}.

\subsection{Concept Estimation from Expert Trajectories}
\label{sec:concept-estimation}
For a new task, let $\mathcal D_{\mathrm{traj}}$ contain its selected expert
trajectories. We construct one decision example for each environment-action
step in these trajectories. At such a step $t$, the input is the
pre-action context $x_t$ defined in Section~\ref{sec:sequential-setup},
the prediction target is the recorded expert action $a_t$, and a finite
candidate collection $\mathcal A_t$ is stored with that step.
Each candidate collection must contain the recorded action.
The context includes only information available before $a_t$, excluding
the action itself and its subsequent feedback.
\texttt{Think} steps do not supply fitting targets, but their reasoning
text and \texttt{OK} feedback remain in the contexts of later decisions.
We then pool the examples across trajectories and index them by
$i=1,\ldots,N$, where $N\geq1$ is the total number of decision examples.
Writing $a_i^{\mathrm E}$ for the expert action in example $i$, the
resulting dataset is
\begin{equation}
 \mathcal D=\{(x_i,\mathcal A_i,a_i^{\mathrm E})\}_{i=1}^{N}.
 \label{eq:decision-data}
\end{equation}
Thus, $\mathcal D_{\mathrm{traj}}$ contains complete trajectories,
whereas $\mathcal D$ contains their environment-action decisions.
The index $t$ denotes a position within an original trajectory, including
\texttt{Think} steps; $i$ enumerates only the pooled fitting examples.

The fixed/open distinction refers to the environment-action space.
For fixed-action-space tasks, the environment supplies environment-action
candidates. For open-action-space tasks, we construct candidates by
sampling multiple replies from the acting model under the same
ReAct context and extracting environment-action strings. This produces
a finite candidate collection without enumerating the full action space.
During expert data collection, the expert selects from the constructed
candidates, which are saved with the trajectory for concept estimation.
Appendix~\ref{app:implementation} provides the candidate-construction
settings and preprocessing details.

We estimate the concept by regularized maximum likelihood.
Let $\lambda>0$ be the regularization strength, $\mathcal L$ the
fitting objective, and $\widehat{\bm\alpha}$ the estimated concept.
With $\|\cdot\|_2$ denoting the Euclidean norm, we minimize the average
negative log-probability of expert actions plus a squared-norm penalty:
\begin{equation}
 \begin{gathered}
 \widehat{\bm\alpha}=\arg\min_{\bm\alpha}\mathcal L(\bm\alpha),\\
 \mathcal L(\bm\alpha)=
 -\frac{1}{N}\sum_{i=1}^{N}
   \log P_{\bm\alpha}(a_i^{\mathrm E}\mid x_i,\mathcal A_i)
 +\frac{\lambda}{2}\|\bm\alpha\|_2^2
 .
 \end{gathered}
 \label{eq:concept-objective}
\end{equation}
Only the concept vector is optimized. Candidate features and prior
weights are computed before fitting and held fixed, along with $\theta$.
The fitted concept can be reused across test tasks that use the same
expert demonstrations, backbone, and feature construction.
Each loss term compares the expert action with its alternatives, and
pooling these terms fits one concept across trajectories. Decisions are
weighted equally, so trajectories with more environment-action decisions
contribute more terms.

With fixed features, positive prior weights, and $\lambda>0$, the
objective in Equation~\eqref{eq:concept-objective} is strongly convex and
has a unique finite minimizer, which we approximate numerically.
Appendix~\ref{app:objective-properties} proves these optimization
properties and states the conditions for interpreting the data term
as a trajectory negative log-likelihood.

\subsection{Online Action Selection}
\label{sec:online-selection}
At each test step, a ReAct response determines whether the agent reasons
or takes an environment action. A \texttt{Think} response is recorded
with fixed feedback \texttt{OK}, updating the history for the next step.
On the environment-action branch, the agent constructs $x_t$ and
$\mathcal A_t$, recomputes candidate features and prior weights, and selects
\begin{equation}
 a_t=\arg\max_{a\in\mathcal A_t}
 P_{\widehat{\bm\alpha}}(a\mid x_t,\mathcal A_t).
 \label{eq:execute-action}
\end{equation}
Ties are resolved by a fixed rule. The selected environment action is
executed and its observation is appended to the history. The concept
$\widehat{\bm\alpha}$ remains fixed throughout the task, and $\theta$
is frozen throughout fitting and execution.

The ReAct prompt includes demonstrations to guide reasoning and routing.
The action scorer instead uses $x_t$, containing the current task and
its interaction history, including preceding thoughts, but no expert
demonstrations. Thus, the fitted concept influences environment-action
selection, while reasoning remains generated by the language model.

Candidate generation determines which actions can be considered;
Equation~\eqref{eq:execute-action} determines which one is selected.
Only the selected action is executed, and unselected proposals are not
tested in the environment. Algorithm~\ref{alg:cb-agent} summarizes
the procedure for nonempty candidate collections; parsing and fallback
details are discussed in Appendix~\ref{app:implementation}.

\begin{figure}[t]
\refstepcounter{cbalgorithm}\label{alg:cb-agent}
\hrule\medskip
\textbf{Algorithm \thecbalgorithm\quad CB-Agent}
\medskip\hrule\medskip
\small

\textbf{Inputs:} Expert trajectories $\mathcal D_{\mathrm{traj}}$ selected for the task;
frozen model $\theta$; regularization $\lambda>0$;
task instruction $\mathrm{inst}$ and environment.

\begin{algorithmic}[1]
\Statex \textbf{(a) Decision-dataset construction}
\For{each environment-action step $t$ in each trajectory of $\mathcal D_{\mathrm{traj}}$}
    \State Reconstruct $x_t$ from the instruction and preceding history, retaining thoughts.
    \State Retrieve the saved candidates $\mathcal A_t$ and expert action $a_t\in\mathcal A_t$.
\EndFor
\State Pool these decisions as $\mathcal D=\{(x_i,\mathcal A_i,a_i^{\mathrm E})\}_{i=1}^{N}$.
\Statex \textbf{(b) Concept estimation}
\State Cache $\bm\phi(x_i,a;\mathcal A_i)$ and $\overline Q_\theta(a\mid x_i)$ for all $i$ and $a\in\mathcal A_i$.
\State Fit $\widehat{\bm\alpha}$ by minimizing Equation~\eqref{eq:concept-objective} with the cached quantities fixed.
\Statex \textbf{(c) Online action selection}
\State Receive $o_0$ (empty if unavailable); initialize the interaction record and $t\gets0$.
\While{the interaction has not terminated and the budget remains}
    \State Generate a ReAct response using demonstrations and current history.
    \If{the response is \texttt{Think}}
        \State Set $a_t$ to the \texttt{Think} response and $o_{t+1}\gets\texttt{OK}$.
    \Else
        \State Construct $x_t$ without demonstrations.
        \State Obtain environment-action candidates $\mathcal A_t$ (Section~\ref{sec:concept-estimation}).
        \State Compute $\bm\phi(x_t,a;\mathcal A_t)$ and $\overline Q_\theta(a\mid x_t)$ for $a\in\mathcal A_t$.
        \State Select $a_t$ by Equation~\eqref{eq:execute-action}; execute it to receive $o_{t+1}$.
    \EndIf
    \State Append $(a_t,o_{t+1})$ to the interaction record; set $t\gets t+1$.
\EndWhile
\end{algorithmic}

\textbf{Fixed during online action selection:}
$\widehat{\bm\alpha}$ and $\theta$.
\medskip\hrule
\end{figure}

\section{Experiments}
\label{sec:experiments}
Section~\ref{sec:experimental-setup} describes the evaluation setting,
followed by cross-benchmark results (Section~\ref{sec:main-results}),
concept visualizations (Section~\ref{sec:concept-visualization}),
combination with Reflexion (Section~\ref{sec:reflexion-experiment}),
and a control comparison for the learned concept
(Section~\ref{sec:concept-control}).

\subsection{Experimental Setup}
\label{sec:experimental-setup}
\paragraph{Benchmarks and metrics.}
We evaluate on TextWorld, a framework for text-based games
\citep{cote2018textworld}; ALFWorld, which expresses household tasks
as text-based interaction \citep{shridhar2021alfworld}; and ScienceWorld,
which requires agents to carry out scientific tasks in an interactive
environment \citep{wang2022scienceworld}.
In our TextWorld and ALFWorld settings, the environment supplies
environment-action candidates. ScienceWorld uses an open-action-space
setting in which the model proposes a finite candidate collection at
each step. We report normalized scores for TextWorld and ScienceWorld
and success rate over 134 tasks for ALFWorld, all on a 0--100 scale.
Avg.\ is the unweighted mean of these three metrics, computed before
rounding. It is a cross-benchmark summary, not an overall success rate.

\paragraph{Models and baselines.}
We use Qwen3-4B, Qwen3-8B, and Qwen3-14B.
ReAct interleaves reasoning and actions, whereas Act-only omits explicit
reasoning \citep{yao2023react}. Self-Consistency provides a comparison
based on aggregating sampled responses \citep{wang2023selfconsistency}.
StateAct makes goals and states explicit \citep{rozanov2025stateact},
while ReflAct reflects on the current state relative to the goal
\citep{kim2025reflact}. ExpeL reuses natural-language insights and retrieved
trajectories \citep{zhao2024expel}. We also include DeltaMem \citep{tan2026deltamem} and a
Trajectory SFT reference that updates model parameters using trajectory
data.
Reflexion configurations with two attempts are evaluated separately.

\paragraph{Demonstrations and concept fitting.}
Qwen3-32B generates the expert trajectories. Each task type uses a fixed
set of two trajectories for ALFWorld and ten for TextWorld and ScienceWorld,
shared across its test tasks and demonstration-based baselines.
The test task's own expert trajectory is excluded from fitting and
parameter selection. For each backbone, CB-Agent fits one concept per
task type using Equation~\eqref{eq:concept-objective} and reuses it without
test-time updates, keeping the language model frozen. ScienceWorld
candidates are sampled by the agent's own backbone.
Appendix~\ref{app:implementation} provides collection, fitting, and
execution details.

\paragraph{Evaluation scope.}
The main comparison evaluates single-trial task performance. We report
the two-trial Reflexion comparison separately. Computational costs and
the scope of the empirical evidence are discussed in
Appendix~\ref{app:limitations}.

\subsection{Main Results}
\label{sec:main-results}
Table~\ref{tab:main-results} shows that CB-Agent achieves the highest
cross-benchmark average among the evaluated methods at every model size:
66.6, 72.5, and 75.9 for 4B, 8B, and 14B, respectively.
It leads in six of nine model--benchmark combinations, and its scores
increase with model size on all three benchmarks.

On TextWorld, CB-Agent ranks first throughout, including at 8B where
it exceeds the strongest baseline, DeltaMem, by 2.1 points.
On ScienceWorld, ExpeL is the strongest baseline at every size, but
CB-Agent consistently surpasses it. Thus, the leading results span
both environment-supplied and model-generated candidate collections.
The experience-based baselines are less consistent across environments:
ExpeL is strongest on ScienceWorld, whereas DeltaMem is competitive
on TextWorld at 8B.
ALFWorld shows a different pattern: StateAct leads at 4B and 8B,
and ReflAct leads at 14B. CB-Agent improves from 76.9 to 81.3
but remains below these leaders. One possible limitation is that better
individual action choices may still leave errors that prevent task
completion. Since ALFWorld measures complete-task success, such partial
improvements would not increase its reported score.

Finally, relative to ReAct, CB-Agent improves eight settings and matches
one, with average gains of 2.4, 5.1, and 2.5 points at the
three model sizes, computed before rounding.

\begin{table}[t]
\centering
\caption{Results across three benchmarks and model sizes.
All metrics use a 0--100 scale: normalized scores for TextWorld and
ScienceWorld, and success rate for ALFWorld.
Avg.\ is the unweighted mean across the three benchmarks.
Bold marks the best result within each model block.}
\label{tab:main-results}
\small
\setlength{\tabcolsep}{5pt}
\begin{tabular}{llrrrr}
\toprule
 & & \multicolumn{2}{c}{Fixed action space} & Open action space & \\
\cmidrule(lr){3-4}\cmidrule(lr){5-5}
Model & Method & TextWorld & ALFWorld & ScienceWorld & Avg. \\
\midrule
\multirow{9}{*}{Qwen3-4B} & ReAct & 61.8 & 76.9 & 53.9 & 64.2 \\
 & Act-only & 52.0 & 54.5 & 48.6 & 51.7 \\
 & Self-Consistency & 57.4 & 69.4 & 35.4 & 54.1 \\
 & StateAct & 43.3 & \textbf{79.1} & 28.7 & 50.4 \\
 & ExpeL & 37.0 & 67.2 & 58.7 & 54.3 \\
 & ReflAct & 57.2 & 50.0 & 28.7 & 45.3 \\
 & DeltaMem & 49.7 & 41.8 & 31.4 & 41.0 \\
 & Trajectory SFT & 26.6 & 22.4 & 4.9 & 18.0 \\
\cmidrule(lr){2-6}
 & CB-Agent (ours) & \textbf{63.5} & 76.9 & \textbf{59.3} & \textbf{66.6} \\
\midrule
\multirow{9}{*}{Qwen3-8B} & ReAct & 74.6 & 72.4 & 55.1 & 67.4 \\
 & Act-only & 62.4 & 68.7 & 36.6 & 55.9 \\
 & Self-Consistency & 54.7 & 79.9 & 44.0 & 59.5 \\
 & StateAct & 38.5 & \textbf{82.8} & 17.6 & 46.3 \\
 & ExpeL & 45.5 & 68.7 & 58.7 & 57.6 \\
 & ReflAct & 61.3 & 56.0 & 35.8 & 51.0 \\
 & DeltaMem & 75.4 & 62.7 & 36.2 & 58.1 \\
 & Trajectory SFT & 49.2 & 11.2 & 8.3 & 22.9 \\
\cmidrule(lr){2-6}
 & CB-Agent (ours) & \textbf{77.5} & 78.4 & \textbf{61.6} & \textbf{72.5} \\
\midrule
\multirow{9}{*}{Qwen3-14B} & ReAct & 80.6 & 78.4 & 61.4 & 73.5 \\
 & Act-only & 75.9 & 74.6 & 59.5 & 70.0 \\
 & Self-Consistency & 57.1 & 81.3 & 50.8 & 63.1 \\
 & StateAct & 46.0 & 73.1 & 26.4 & 48.5 \\
 & ExpeL & 39.1 & 79.9 & 63.5 & 60.8 \\
 & ReflAct & 72.6 & \textbf{84.3} & 46.1 & 67.7 \\
 & DeltaMem & 67.0 & 74.6 & 49.4 & 63.7 \\
 & Trajectory SFT & 52.9 & 32.1 & 11.3 & 32.1 \\
\cmidrule(lr){2-6}
 & CB-Agent (ours) & \textbf{82.3} & 81.3 & \textbf{64.1} & \textbf{75.9} \\
\bottomrule
\end{tabular}
\end{table}

\subsection{Visualization of Estimated Concepts}
\label{sec:concept-visualization}
\begin{figure}[t]
\centering
\begin{minipage}[t]{0.325\textwidth}
 \centering
 \includegraphics[width=\linewidth]{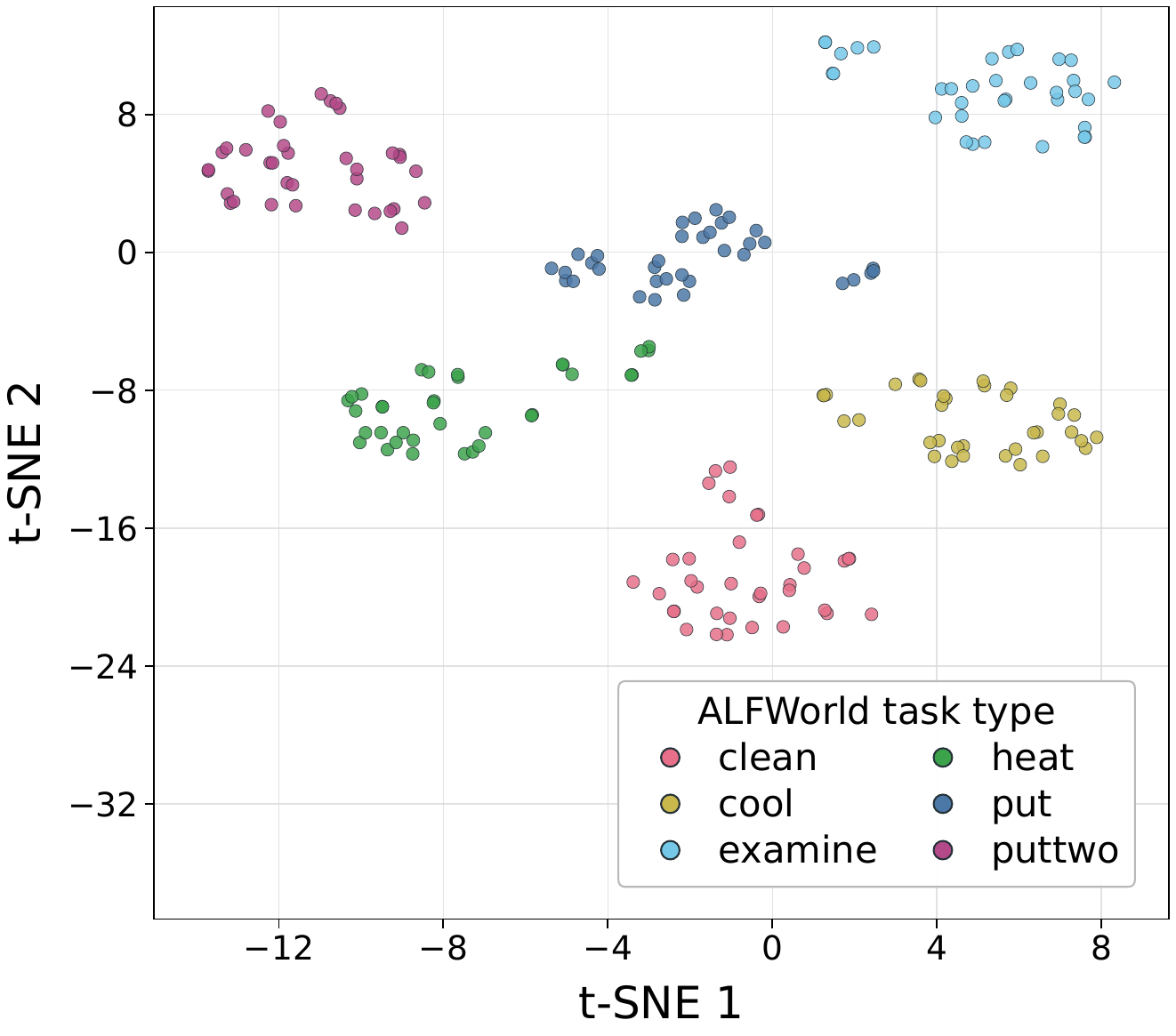}\\
 \small (a) Qwen3-4B
\end{minipage}\hfill
\begin{minipage}[t]{0.325\textwidth}
 \centering
 \includegraphics[width=\linewidth]{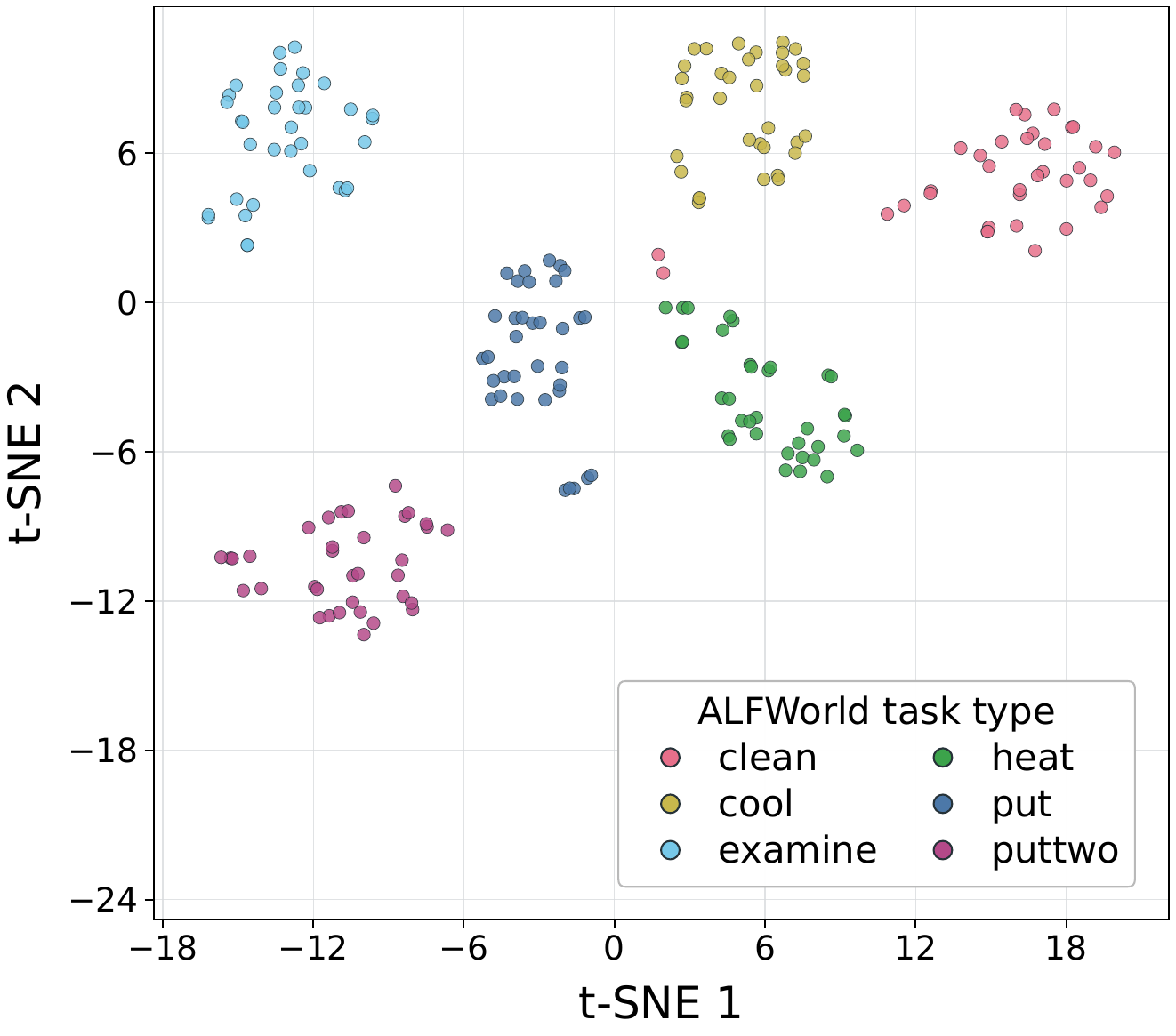}\\
 \small (b) Qwen3-8B
\end{minipage}\hfill
\begin{minipage}[t]{0.325\textwidth}
 \centering
 \includegraphics[width=\linewidth]{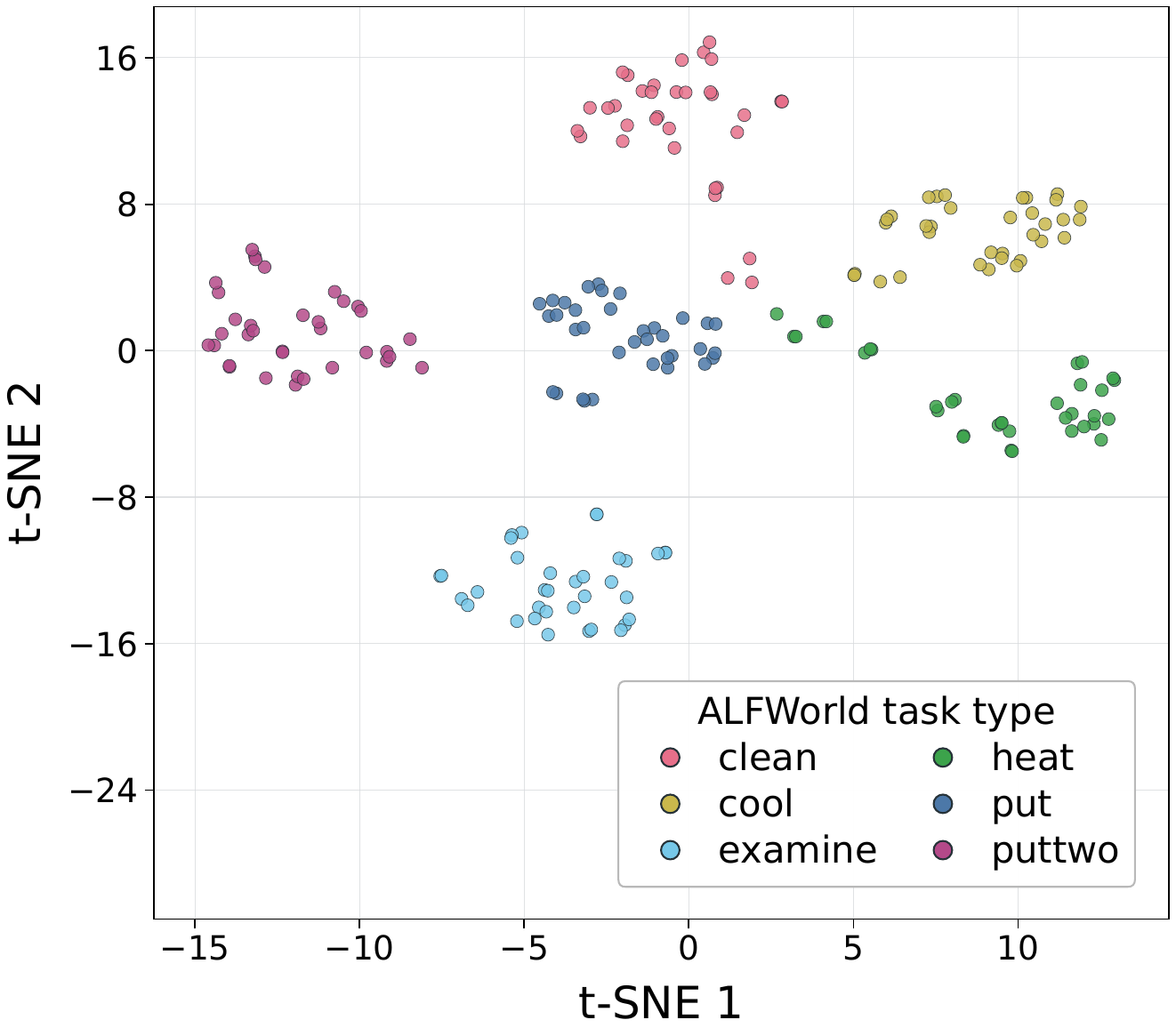}\\
 \small (c) Qwen3-14B
\end{minipage}
\caption{t-SNE visualizations of concept vectors on ALFWorld for (a) Qwen3-4B, (b) Qwen3-8B, and (c) Qwen3-14B. Each point represents a concept estimated from five randomly sampled expert trajectories of a single task type, with colors denoting the six task types. Concepts from the same task type tend to cluster across all three backbones. Each panel is projected independently.}
\label{fig:concept-tsne}
\end{figure}

For each of ALFWorld's six task types, we repeatedly sample five complete
expert trajectories at random and fit one concept per set using
Section~\ref{sec:concept-estimation}. We apply t-SNE
\citep{vandermaaten2008tsne} separately for each backbone.
This diagnostic uses five trajectories per concept, rather than the
two used in the main ALFWorld evaluation.
Figure~\ref{fig:concept-tsne} shows recurring same-type clusters across
all three backbones, with some dispersion and outliers. This provides qualitative evidence of task-related structure across
sampled trajectory sets, suggesting that the estimated concepts capture
patterns shared by tasks of the same type.

\subsection{Combining CB-Agent with Reflexion}
\label{sec:reflexion-experiment}
We evaluate whether concept-based action selection can be combined with
Reflexion, which uses feedback from earlier attempts to guide subsequent
ones \citep{shinn2023reflexion}.
On ALFWorld, we compare Reflexion with CB-Agent + Reflexion under a
two-trial setting for each backbone. The concept remains fixed:
reflection changes the interaction context rather than refitting the
concept from test feedback. This comparison is separate from the main
table because additional attempts introduce a different evaluation budget.
Table~\ref{tab:alfworld-reflexion} shows success-rate gains of 4.5, 7.5,
and 5.2 percentage points at 4B, 8B, and 14B, respectively.
These results support combining a fixed concept-based scoring rule
with reflections that update the context across attempts.

\subsection{Effect of the Learned Concept}
\label{sec:concept-control}
To isolate the contribution of the fitted concept, we compare CB-Agent
with a control that sets $\bm\alpha=\bm 0$ on ALFWorld using all three
backbones. Both configurations retain the same ReAct routing, candidate
construction, language-model prior, length correction, and interaction
budget. With the concept term removed, the control selects actions using
only the length-adjusted prior weights in
Equation~\eqref{eq:conditional-action-model}; it is neither random
selection nor standard ReAct. Table~\ref{tab:concept-control} reports
success rates, with CB-Agent values reproduced from
Table~\ref{tab:main-results}.
The learned concept improves success rates by 10.4, 6.0, and 14.9
percentage points at 4B, 8B, and 14B, respectively, computed before
rounding. Gains across all three backbones support its contribution
beyond the length-adjusted language-model prior in this ALFWorld setting.

These gains address a different question from the comparison with ReAct.
The main table compares complete agent configurations, whereas this
control tests whether learning from expert choices improves the same
candidate-scoring pipeline. Although CB-Agent does not lead on ALFWorld,
its gains over prior-only selection across all three backbones support
the contribution of the learned concept within this pipeline. Together
with the Reflexion results, this supports using a fixed concept as a
reusable component: expert choices inform the scoring rule, while
interaction feedback and reflections change the context in which it is applied.

\begin{table}[t]
\centering
\begin{minipage}[t]{0.49\textwidth}
\centering
\caption{ALFWorld success rate (\%) over 134 tasks with Qwen3 backbones.
Both methods use two trials. Bold marks the higher value.}
\label{tab:alfworld-reflexion}
\small
\setlength{\tabcolsep}{3pt}
\begin{tabular}{@{}lccc@{}}
\toprule
Method & 4B & 8B & 14B \\
\midrule
Reflexion & 85.1 & 82.1 & 85.8 \\
CB-Agent + Reflexion & \textbf{89.6} & \textbf{89.6} & \textbf{91.0} \\
\bottomrule
\end{tabular}
\end{minipage}\hfill
\begin{minipage}[t]{0.49\textwidth}
\centering
\caption{ALFWorld success rate (\%) over 134 tasks with Qwen3 backbones.
Single-trial concept control. Bold marks the higher value.}
\label{tab:concept-control}
\small
\setlength{\tabcolsep}{3pt}
\begin{tabular}{@{}lccc@{}}
\toprule
Configuration & 4B & 8B & 14B \\
\midrule
$\bm\alpha=\bm 0$ & 66.4 & 72.4 & 66.4 \\
CB-Agent & \textbf{76.9} & \textbf{78.4} & \textbf{81.3} \\
\bottomrule
\end{tabular}
\end{minipage}
\end{table}

\section{Related Work}
\label{sec:related-work}
ICL research explores numerical representations of demonstration
knowledge, including Task Vectors and Function Vectors
\citep{hendel2023taskvectors,todd2024functionvectors}.
CB-ICL estimates shared concepts for individual-input prediction
\citep{tang2025theoretical}; we adapt this framework to sequential
action selection with changing contexts and candidates.
For language agents, ReAct interleaves reasoning and actions
\citep{yao2023react}, while ExpeL reuses language-based insights and
trajectories \citep{zhao2024expel}, and Reflexion uses verbal feedback
across attempts \citep{shinn2023reflexion}.
CB-Agent retains ReAct interaction but fits a numerical concept from
expert choices to score actions, without requiring verbal experience
summaries or updating LLM parameters.
Appendix~\ref{app:extended-related-work} provides an extended discussion.

\section{Conclusion}
We introduced CB-Agent, which adapts concept-based in-context learning
to sequential action selection. It fits a reusable concept from expert
environment-action decisions and uses frozen language-model features
to score candidates in the evolving interaction context, without first
converting experience into verbal rules. Across three benchmarks and
three Qwen3 model sizes, CB-Agent achieves the highest average score
among the evaluated methods. An ALFWorld control comparison supports
the contribution of the learned concept beyond prior-only action scoring.
Combining CB-Agent with Reflexion further improves ALFWorld success rates,
while concept visualizations provide qualitative evidence of task-related
structure. Future work will examine candidate coverage and transfer to
unseen task types.

\clearpage
\subsection*{AI use statement}
In this work, we used OpenAI Codex to assist with interpreting author-provided experimental results and improve the language and readability of the manuscript. We have not used generative AI tools to develop theoretical models, formulate mathematical claims, write proofs, propose hypotheses, or implement the proposed method.
As part of the experimental workflow, we used Qwen3-32B to generate
the expert demonstration trajectories used in this study. The roles
of language models in demonstration generation and agent execution
are described in the method and experimental sections.
The authors retain responsibility for all research decisions and
for the accuracy, originality, and integrity of the final manuscript,
including its mathematical claims, experimental interpretations,
references, and AI-assisted artifacts.

\subsection*{Ethics statement}
This work evaluates language agents in existing benchmarks: TextWorld,
ALFWorld, and ScienceWorld. The experiments are conducted in simulated
environments and do not involve human participants, personal data
collection, or real-world deployment.

\subsection*{Reproducibility statement}
Section~\ref{sec:method} specifies the action model, concept-fitting
objective, and interaction procedure.
Section~\ref{sec:experimental-setup} describes the benchmarks, metrics,
backbones, and shared-demonstration protocol.
Appendix~\ref{app:implementation} documents feature construction,
length-adjusted prior weights, trajectory processing, optimization,
and candidate generation. Appendix~\ref{app:objective-properties}
provides derivations of the objective's properties under the stated
assumptions. The source code is available at \url{https://anonymous.4open.science/r/CB-Agent-DFF9/}.

\bibliography{iclr2027_conference}
\bibliographystyle{iclr2027_conference}

\appendix
\clearpage
\section*{Appendix Contents}
\begingroup
\small
\renewcommand{\arraystretch}{1.25}
\begin{tabular}{@{}p{0.88\linewidth}r@{}}
\hyperref[app:notation]{\ref*{app:notation}\quad\nameref*{app:notation}} & \pageref{app:notation} \\
\hyperref[app:implementation]{\ref*{app:implementation}\quad\nameref*{app:implementation}} & \pageref{app:implementation} \\
\hyperref[app:objective-properties]{\ref*{app:objective-properties}\quad\nameref*{app:objective-properties}} & \pageref{app:objective-properties} \\
\hyperref[app:exponential-motivation]{\ref*{app:exponential-motivation}\quad\nameref*{app:exponential-motivation}} & \pageref{app:exponential-motivation} \\
\hyperref[app:extended-related-work]{\ref*{app:extended-related-work}\quad\nameref*{app:extended-related-work}} & \pageref{app:extended-related-work} \\
\hyperref[app:limitations]{\ref*{app:limitations}\quad\nameref*{app:limitations}} & \pageref{app:limitations} \\
\end{tabular}
\endgroup

\newpage

\section{Notation}
\label{app:notation}
Table~\ref{tab:notation} summarizes the principal notation used in the
paper. The trajectory index $t$ and pooled-example index $i$ refer to
different units: $t$ includes reasoning steps, whereas $i$ indexes only
environment-action decisions used for fitting.

\begingroup
\small
\renewcommand{\arraystretch}{1.12}
\begin{longtable}{@{}p{0.28\textwidth}p{\dimexpr0.72\textwidth-2\tabcolsep\relax}@{}}
\caption{Principal notation for CB-ICL and CB-Agent.}\label{tab:notation}\\
\toprule
Symbol & Meaning \\
\midrule
\endfirsthead
\multicolumn{2}{l}{\tablename~\thetable\ (continued)}\\
\toprule
Symbol & Meaning \\
\midrule
\endhead
\midrule
\multicolumn{2}{r}{Continued on the next page}\\
\endfoot
\bottomrule
\endlastfoot
\multicolumn{2}{@{}l}{\textbf{Prediction and interaction}}\\
$x$, $y$, $\mathcal Y$ & Input text, candidate label, and finite nonempty label set in CB-ICL.\\
$Q_0(y)$ & Positive base weight for label $y$ in the CB-ICL model.\\
$\mathrm{inst}$ & Task instruction specifying the goal of an interaction.\\
$\tau$, $T$ & Complete interaction trajectory and its number of steps, including both environment and \texttt{Think} actions.\\
$t$, $a_t$, $o_{t+1}$ & Interaction-step index, environment or \texttt{Think} action, and recorded feedback.\\
$o_0$ & Initial observation before any action; empty if the environment provides none.\\
$x_t$ & Pre-action text context containing the instruction, initial observation, and preceding actions and feedback.\\
$\mathcal A_t$ & Finite nonempty candidate collection at environment-action step $t$.\\
$x$, $a$, $\mathcal A$ & Context, candidate environment action, and candidate collection for one decision, with its index omitted.\\
\midrule
\multicolumn{2}{@{}l}{\textbf{Expert data and concept estimation}}\\
$\mathcal D_{\mathrm{traj}}$ & Collection of selected complete expert trajectories.\\
$\mathcal D$, $N$ & Pooled dataset of expert environment-action decisions and its number of examples.\\
$i$, $(x_i,\mathcal A_i,a_i^{\mathrm E})$ & Pooled-example index and its context, saved candidates, and recorded expert environment action.\\
$\theta$ & Frozen language-model parameters.\\
$m$, $\bm\phi(x,a;\mathcal A)$ & Feature dimension and fixed action-feature vector in $\mathbb R^m$ for a given context and candidate collection.\\
$\bm\alpha$, $\widehat{\bm\alpha}$ & Concept parameter vector in $\mathbb R^m$ and its fitted estimate.\\
$P_{\bm\alpha}$ & Concept-conditioned distribution over candidate labels or environment actions, as specified by its arguments.\\
$\mathcal L$, $\lambda$ & Regularized negative log-likelihood objective and positive regularization strength.\\
\midrule
\multicolumn{2}{@{}l}{\textbf{Language-model weights and feature construction}}\\
$Q_\theta(a\mid x)$ & Frozen model probability of the complete candidate action's token sequence.\\
$L(a)$, $\gamma$ & Number of action tokens and length-correction exponent in $[0,1]$.\\
$\overline Q_\theta(a\mid x)$ & Length-adjusted prior weight, $Q_\theta(a\mid x)^{1/L(a)^\gamma}$.\\
$p_\theta$, $a_\ell$, $a_{<\ell}$ & Next-token distribution, action token at position $\ell$, and preceding action tokens.\\
$d$, $\bm u_{i,a}$ & Hidden width and last-layer hidden vector at the final token of candidate $a$ in example $i$; $m=d+1$ in the implemented feature construction.\\
$M_i$ & Number of candidate entries, $|\mathcal A_i|$.\\
$\mu_{i,j}$, $\sigma_{i,j}$ & Mean and standard deviation of hidden coordinate $j$ across candidates in example $i$.\\
$\widetilde{\bm u}_{i,a}$, $\epsilon$ & Coordinate-wise standardized hidden vector and positive numerical floor used when dividing by standard deviations.\\
\end{longtable}
\endgroup

\section{Representation and Execution Details}
\label{app:implementation}

\paragraph{Candidate features.}
Let $d$ be the language model's hidden width, so that $m=d+1$ in our
feature construction. For decision $i$ and candidate $a\in\mathcal A_i$,
let $\bm u_{i,a}\in\mathbb R^d$ be the last-layer hidden vector at the
final token of $a$, obtained by encoding the context followed by the
action prefix and candidate string. The ALFWorld implementation
uses plain-text serialization with the prefix \texttt{\textbackslash n> },
rather than applying a chat template. Let $M_i=|\mathcal A_i|$ be the number
of candidate entries. For coordinate $j\in\{1,\ldots,d\}$, let $u_{i,a,j}$
be coordinate $j$ of $\bm u_{i,a}$, and let $\mu_{i,j}$ and $\sigma_{i,j}$
be its mean and standard deviation across candidates. Choose a numerical
floor $\epsilon>0$ to avoid division by zero. Denote the standardized coordinate by
$\widetilde u_{i,a,j}$ and the vector collecting these coordinates by
$\widetilde{\bm u}_{i,a}$:
\begin{align}
 \mu_{i,j}&=\frac{1}{M_i}\sum_{a\in\mathcal A_i}u_{i,a,j},
 &\sigma_{i,j}&=
 \sqrt{\frac{1}{M_i}\sum_{a\in\mathcal A_i}(u_{i,a,j}-\mu_{i,j})^2},
 \\
 \widetilde u_{i,a,j}&=
 \frac{u_{i,a,j}-\mu_{i,j}}{\max(\sigma_{i,j},\epsilon)},
 &\bm\phi(x_i,a;\mathcal A_i)&=
 \begin{bmatrix}\widetilde{\bm u}_{i,a}\\M_i^{-1/2}\end{bmatrix}.
 \label{eq:candidate-feature}
\end{align}
We recompute the means and standard deviations for each candidate
collection, including at test time, without using expert labels.
The appended coordinate $M_i^{-1/2}$ is a normalized bias feature
shared by all candidates at decision $i$, following the feature
construction in CB-ICL \citep{tang2025theoretical}.

\paragraph{Language-model weights.}
For token index $\ell\in\{1,\ldots,L(a)\}$, let $a_\ell$ be token $\ell$
of candidate action $a$ and $a_{<\ell}$ its preceding tokens.
Let $p_\theta$ be the frozen model's next-token distribution. The model
probability $Q_\theta(a\mid x)$ is the product of these token probabilities,
so its length-adjusted log-weight is
\begin{equation}
 \log\overline Q_\theta(a\mid x)=
 \frac{1}{L(a)^\gamma}\sum_{\ell=1}^{L(a)}
 \log p_\theta(a_\ell\mid x,a_{<\ell}).
 \label{eq:length-normalized-prior}
\end{equation}
When computing token probabilities, the input includes the model's reply
prefix. The length $L(a)$ counts only action tokens, not context or
role-prefix tokens. Fitting and execution use the same text formatting
and token-counting convention.
All candidates are assumed to have positive model probability.
The reported implementation first normalizes these prior weights over
candidates, then adds their logarithms to the concept scores.
Ignoring numerical stabilization, prior normalization divides every weight by the same number, which
cancels between the numerator and denominator of
Equation~\eqref{eq:conditional-action-model}. Using the unnormalized
weights therefore gives the same action distribution.

\paragraph{Think actions and trajectory construction.}
A \texttt{Think} action contains the reasoning text as part of its action
string and receives the fixed feedback \texttt{OK}. The ReAct response
determines the branch: a response beginning with \texttt{think:} is
recorded directly; otherwise, the concept scorer selects an environment
action from its candidates. The initial action proposed by the routing
response is not the executed action unless the scorer selects it.
\texttt{Think} responses contribute no likelihood terms. Their text and
feedback are included only in subsequent contexts. A \texttt{Think}
step advances the interaction index without changing the environment.

The expert Qwen3-32B model uses native thinking, but content inside its
\texttt{<think>} tags is discarded. The stored reasoning is the explicit
agent-level \texttt{Think} response outside those tags. The evaluated
4B, 8B, and 14B agents use explicit \texttt{Think} responses without native
model thinking.

\paragraph{Demonstration selection and fitting scope.}
As specified in Section~\ref{sec:experimental-setup}, demonstrations are
fixed by task type, and each backbone's fitted concept is reused across
test tasks of that type. It also remains fixed during reflection-based
retries. Test-time feedback updates the interaction context but is not
used to refit the concept.
During expert collection, candidate actions and the expert's choice are
saved together. Each evaluated backbone recomputes features and prior
weights on these saved contexts and candidates. Only environment-action
steps contribute fitting targets; preceding \texttt{Think} responses
remain in their contexts.

\paragraph{Optimization and execution settings.}
We use the following settings across TextWorld, ALFWorld, and ScienceWorld.
We initialize the concept at zero and use
Adam with learning rate $0.01$, at most 1,500 iterations, and gradient-norm
tolerance $10^{-4}$. Its squared-norm penalty has coefficient $0.5$,
corresponding to $\lambda=1$ in Equation~\eqref{eq:concept-objective}.
Candidate features and prior weights are cached before fitting.
Each trial allows at most 50 interaction steps, including
\texttt{Think} steps. We filter \texttt{look} and \texttt{inventory}
from the environment-action candidates, with \texttt{look} as the
empty-collection fallback.
We use length exponent $\gamma=0.8$
and unit weight on the prior log-probability.

\paragraph{Open-action-space candidate construction.}
ScienceWorld requests 20 independently sampled replies to the same
ReAct context at temperature $0.7$, rather than asking for a list of
20 actions in one reply. Replies may repeat, so this does not guarantee
20 distinct candidates. Each retained candidate is a single environment-action
string; \texttt{Think} is handled by routing, not by the concept scorer.
During expert collection, Qwen3-32B proposes and selects candidates.
During evaluation, the agent's own backbone proposes candidates.
Fitting reuses the saved expert candidate collections and recomputes
their features and prior weights with the evaluated backbone.
The recorded expert target is therefore already among its saved candidates.
Only the selected environment action is executed, and candidates are
resampled at the next environment-action decision.

\section{Properties of the Conditional Objective}
\label{app:objective-properties}

This appendix establishes properties of the fitting objective for a
given dataset, with features and prior weights held fixed. The analysis
concerns the optimization objective; predictive accuracy and task
performance are evaluated experimentally.

\paragraph{Existence and uniqueness.}
Let $\nabla^2\mathcal L$ denote the Hessian of the objective with respect
to $\bm\alpha$, $I_m$ the $m$-dimensional identity matrix, and
$\operatorname{Cov}$ the covariance under the indicated candidate
distribution. With fixed features and positive prior weights,
\begin{equation}
 \nabla^2\mathcal L(\bm\alpha)=
 \frac{1}{N}\sum_{i=1}^{N}
 \operatorname{Cov}_{a\sim P_{\bm\alpha}(\cdot\mid x_i,\mathcal A_i)}
 [\bm\phi(x_i,a;\mathcal A_i)]+\lambda I_m.
\end{equation}
Each covariance is positive semidefinite, so the objective is
$\lambda$-strongly convex. The negative log-likelihood is nonnegative,
and the quadratic penalty makes the objective grow without bound as
$\|\bm\alpha\|_2$ grows. Consequently,
a unique finite minimizer exists.


\paragraph{Trajectory interpretation.}
The objective weights expert environment-action decisions equally;
a trajectory with more such decisions contributes more terms.
No independence assumption between consecutive
decisions is needed to define this empirical objective.
To interpret the data loss as a trajectory likelihood, assume that the
initial conditions do not depend on $\bm\alpha$. Also assume that,
given the preceding interaction record and selected action, feedback
does not depend on $\bm\alpha$. Candidate generation and any external
stopping rule must also be independent of $\bm\alpha$ conditional on
the preceding record. The same condition applies to ReAct routing and
reasoning generation. A \texttt{Think} response and its deterministic
\texttt{OK} feedback then contribute only factors independent of
$\bm\alpha$.
Under the action model, the joint probability of a recorded trajectory
and its saved candidate collections factors into the conditional
environment-action probabilities and terms independent of $\bm\alpha$.
Taking negative logarithms, summing over trajectories,
and dividing by $N$ gives the data term in
Equation~\eqref{eq:concept-objective}, up to an additive term independent
of $\bm\alpha$. This interpretation does not require
assuming that the expert policy is exactly realizable by the fitted model.

\section{From Weak Dependence to the Exponential Concept Model}
\label{app:exponential-motivation}

The exponential form in Equation~\eqref{eq:cbicl-exponential} is motivated
by a local representation of conditional distributions. We give the
derivation using the modal decomposition and weak-dependence framework
of \citet{huang2024universal}, then connect it to the concept
parameterization in Section~\ref{sec:cbicl-background}.

\paragraph{Weak dependence as a small conditional correction.}
For this local analysis, let $X$ and $Y$ be random variables on fixed
finite sets $\mathcal X_0$ (inputs) and $\mathcal Y$ (labels), each with
at least two elements. Write $P_{XY}$ for their joint distribution and
$P_X,P_Y$ for its marginals. Consider a family of such distributions
indexed by $\delta>0$, with marginal probabilities bounded below by a
constant $c_0>0$ as $\delta\to0$. We suppress this dependence in the
notation. The weak-dependence condition is
\begin{equation}
 \chi^2(P_{XY}\|P_XP_Y)
 :=\sum_{x\in\mathcal X_0}\sum_{y\in\mathcal Y}
 \frac{\bigl(P_{XY}(x,y)-P_X(x)P_Y(y)\bigr)^2}
 {P_X(x)P_Y(y)}\le\delta^2.
 \label{eq:weak-dependence-appendix}
\end{equation}
Thus, weak dependence concerns the relation between the input $X$ and
label $Y$. Define their relative departure from independence by
\begin{equation}
 r(x,y):=\frac{P_{XY}(x,y)}{P_X(x)P_Y(y)}-1,
 \qquad
 P(y\mid x)=P_Y(y)\bigl[1+r(x,y)\bigr].
 \label{eq:relative-dependence}
\end{equation}
Equation~\eqref{eq:weak-dependence-appendix} implies
$|r(x,y)|\le\delta/\sqrt{P_X(x)P_Y(y)}\le\delta/c_0$.
Also, $\sum_y P_Y(y)r(x,y)=0$ for every $x$ by conditional normalization.
All order statements below are uniform over $x,y$ as $\delta\to0$.

\paragraph{From the linear correction to an exponential form.}
Let $K=\min\{|\mathcal X_0|,|\mathcal Y|\}$.
The modal decomposition supplies functions
$f_j:\mathcal X_0\to\mathbb R$ and $g_j:\mathcal Y\to\mathbb R$,
and coefficients $\sigma_j\ge0$, for $j=1,\ldots,K-1$, such that
\begin{equation}
 r(x,y)=\sum_{j=1}^{K-1}\sigma_j f_j(x)g_j(y).
 \label{eq:modal-correction}
\end{equation}
The functions are mean-zero and orthonormal under their respective
marginals; the coefficients are the nonconstant singular values of the
matrix with entries $P_{XY}(x,y)/\sqrt{P_X(x)P_Y(y)}$
\citep[Proposition~2.2]{huang2024universal}.
Since $r=O(\delta)$, Taylor expansion of $\log(1+r)$ gives
\begin{equation}
 \log\frac{P(y\mid x)}{P_Y(y)}
 =\sum_{j=1}^{K-1}\sigma_j f_j(x)g_j(y)+O(\delta^2).
 \label{eq:local-log-correction}
\end{equation}
This is the local exponential representation: to first order, the same
features that describe a linear change in probability also describe an
additive change in log-probability. The second-order remainder here
follows directly from the stated finite-support and marginal lower-bound
conditions.

\paragraph{Normalization and the concept parameterization.}
To express this representation in the notation of Equation~\eqref{eq:cbicl-exponential},
take $m=K-1$, set $\phi_j(x,y)=f_j(x)g_j(y)$, and collect the modal
coefficients in $\bm\alpha_\star=(\sigma_1,\ldots,\sigma_m)^\top$.
Then $\bm\alpha_\star^\top\bm\phi(x,y)=r(x,y)$.
Define the normalizing factor
$Z_\star(x)=\sum_{b\in\mathcal Y}P_Y(b)\exp\{r(x,b)\}$.
The zero-mean identity above and the expansion
$e^r=1+r+O(\delta^2)$ imply $Z_\star(x)=1+O(\delta^2)$.
Consequently, the normalized exponential distribution satisfies
\begin{equation}
 \frac{P_Y(y)\exp\{\bm\alpha_\star^\top\bm\phi(x,y)\}}
 {\sum_{b\in\mathcal Y}P_Y(b)
 \exp\{\bm\alpha_\star^\top\bm\phi(x,b)\}}
 =P(y\mid x)+O(\delta^2).
 \label{eq:normalized-local-model}
\end{equation}
This gives Equation~\eqref{eq:cbicl-exponential} with base weight
$Q_0=P_Y$ as a first-order accurate model. Normalization ensures a
positive distribution whose probabilities sum to one for any finite
concept vector, including outside the local regime.

The concept model retains this normalized form while using a chosen
positive base weight $Q_0$ and fixed language-model features in place
of the distribution-dependent modal features. These choices define
the model family; demonstration likelihood determines its concept
vector. The $O(\delta^2)$ approximation above applies to the full modal
representation, while the practical model's approximation quality
depends on its chosen features and base weights.
Section~\ref{sec:method} applies this construction to decisions with
changing contexts and candidate actions, using the length-adjusted
language-model weights as the context-dependent baseline.

\section{Extended Related Work}
\label{app:extended-related-work}

\subsection{In-Context Learning and Numerical Task Representations}
ICL uses demonstrations to specify a task without updating the language
model's parameters \citep{brown2020language,dong2024survey}. One line of work improves
the selection and ordering of examples in the prompt
\citep{rubin2022retrieve,lu2022ordered}. For instance,
\citet{liu2022goodexamples} retrieve semantically similar demonstrations
for each input. Another line investigates how a model infers or represents the
task conveyed by demonstrations \citep{xie2022bayesian}.
\citet{hendel2023taskvectors} show that
ICL behavior can often be represented using a task vector computed from
examples. \citet{todd2024functionvectors} identify activation vectors
that encode task functions and can elicit corresponding behavior when
inserted into a model's forward pass. These two studies motivate numerical
representations of demonstration knowledge, but their activation-based
mechanisms differ from fitting weights for an explicit candidate-scoring
model.

CB-ICL connects concept estimation over language-model features with
theoretical analyses of label prediction \citep{tang2025theoretical}.
This framework is the starting point of CB-Agent. We use expert choices
from multiple trajectories to fit a shared concept, then apply it to
context-dependent action features during execution. The key change in
setting is that an action affects the feedback and thus the context of
later decisions; the inputs are not simply a fixed collection of
prediction queries. Candidate collections can also differ from one
step to the next. Accordingly, the prediction analysis of prior CB-ICL
does not by itself establish a task-success guarantee for our agent.
Our objective analysis concerns fitting with fixed features and prior
weights, while the sequential behavior is assessed experimentally.

\subsection{Reasoning, State Tracking, and Action Selection}
Chain-of-thought prompting elicits intermediate reasoning
\citep{wei2022chain}, while tree-based search explores alternative
reasoning paths \citep{yao2023tree}.
ReAct provides a framework for alternating language-based reasoning
with actions and environmental observations \citep{yao2023react}.
Subsequent methods change how this interaction is organized.
StateAct uses self-prompting and state tracking to maintain task goals
and relevant state information \citep{rozanov2025stateact}.
ReflAct emphasizes repeated reflection on the relation between the
current state and the goal \citep{kim2025reflact}.
AdaPlanner revises plans using environmental feedback and also reuses
successful plans as few-shot exemplars \citep{sun2023adaplanner}.
These are overlapping approaches to improving decisions, rather than
mutually exclusive categories: a method can organize the current
interaction while also retaining useful experience.

Self-Consistency offers a different approach to inference: it samples
multiple reasoning paths and selects an answer by aggregating their
outcomes \citep{wang2023selfconsistency}. CB-Agent's open-action-space
procedure also uses multiple sampled replies, but sampling serves to
propose candidates for the current decision. Selection uses a concept
fitted from expert choices together with language-model prior weights,
rather than agreement among sampled answers. Only the selected action
is executed. Thus, candidate proposal and candidate selection are
distinct parts of our procedure; the proposal mechanism is not a
standard component of ReAct. CB-Agent uses ReAct contexts without
introducing a separate planning or state-tracking architecture.

\subsection{Learning and Reusing Agent Experience}
ExpeL and Reflexion show how agents can reuse execution experience
through language. ExpeL extracts insights from training-task experience
and retrieves relevant past trajectories for subsequent decisions
\citep{zhao2024expel}. Reflexion converts feedback into verbal
reflections retained in episodic memory to guide later trials
\citep{shinn2023reflexion}. Their common use of text should not obscure
the difference between transferring experience across tasks and
adapting across repeated attempts at a task. Nor does freezing the LLM
distinguish CB-Agent from these approaches, since they also avoid
updating its parameters.
Other work reuses executable skills or synthesizes task and state
knowledge from trajectories \citep{wang2024voyager,qiao2024worldknowledge}.

CB-Agent differs in what is extracted from experience and how it enters
the decision rule. Expert action choices supervise a numerical concept
through a likelihood objective, and that concept weights action features
when comparing candidates. The method does not require an intermediate
natural-language summary of the expert experience. In contrast to
trajectory-based LLM fine-tuning \citep{zeng2024agenttuning,chen2024agentflan},
the optimized parameters belong to the concept model,
not the LLM. This distinction is about the representation and use of
experience, not a general claim of superiority over language-based
memory or fine-tuning. The concept remains fixed within each test task, while
new observations and any retained reflections change the context used
for scoring. This permits the combination with Reflexion studied in
Section~\ref{sec:reflexion-experiment}, without refitting the concept
from test-time feedback.

\section{Limitations}
\label{app:limitations}

\paragraph{Modeling and transfer.}
CB-Agent models expert action preferences using frozen language-model
features. Our theoretical analysis characterizes the concept-fitting
objective, while end-to-end task performance is evaluated empirically.
The current setting uses demonstrations from the same task type;
performance may depend on their quality and coverage. Transfer to
unseen task types remains an important direction for future work.

\paragraph{Candidate selection and computation.}
In open-action-space tasks, action selection depends on the coverage
and quality of generated candidates. Candidate generation, scoring,
and concept estimation introduce additional computation, and our
comparisons assess task performance rather than efficiency under
matched computational budgets. Demonstrations are omitted from the
action-scoring context but remain in ReAct reasoning and routing,
so the complete agent remains subject to context-window limits.

\paragraph{Evaluation scope.}
Our experiments cover three text-interaction benchmarks and three
model sizes within the Qwen3 family. The prior-only control provides
evidence for the learned concept's contribution on ALFWorld; extending
this analysis to the other benchmarks would clarify its broader
applicability. Evaluation across additional model families and tasks,
together with repeated-run uncertainty estimates, would further
characterize the robustness of the observed gains.

\end{document}